\documentclass[10pt,twocolumn,letterpaper]{article}

\usepackage{iccv}              

\usepackage{times}
\usepackage{epsfig}
\usepackage{graphicx}
\usepackage{amsmath}
\usepackage{amssymb}
\usepackage{pifont}
\usepackage[dvipsnames]{xcolor}

\usepackage{multirow}
\usepackage{multicol}
\usepackage{booktabs}
\usepackage{fontawesome}

\definecolor{iccvblue}{rgb}{0.21,0.49,0.74}
\usepackage[pagebackref,breaklinks,colorlinks,allcolors=iccvblue]{hyperref}

\def\paperID{13520} 
\def\confName{ICCV}
\def\confYear{2025}

\title{VSRM: A Robust Mamba-Based Framework for Video Super-Resolution}

\author{Dinh Phu Tran\textsuperscript{*}\hspace{1cm}
Dao Duy Hung\textsuperscript{*,\textdagger}\hspace{1cm}
Daeyoung Kim\textsuperscript{\ddag}\\[1ex]
{\tt\small \{phutx2000, hicehehe, kimd\}@kaist.ac.kr} \\[1ex]
School of Computing, KAIST, Republic of Korea \\[1ex]
\textsuperscript{*} Equal Contribution\\
\ddag Corresponding author\\
\textdagger Work done during the master program at KAIST
}

\begin{document}
\maketitle
\begin{abstract}
Video super-resolution remains a major challenge in low-level vision tasks.
To date, CNN- and Transformer-based methods have delivered impressive results. However, CNNs are limited by local receptive fields, while Transformers struggle with quadratic complexity, posing challenges for processing long sequences in VSR. Recently, Mamba has drawn attention for its long-sequence modeling, linear complexity, and large receptive fields.
In this work, we propose VSRM, a novel \textbf{V}ideo \textbf{S}uper-\textbf{R}esolution framework that leverages the power of \textbf{M}amba. VSRM introduces Spatial-to-Temporal Mamba 
and Temporal-to-Spatial Mamba blocks 
to extract long-range spatio-temporal features and enhance receptive fields efficiently. To better align adjacent frames, we propose Deformable Cross-Mamba Alignment module. This module utilizes a deformable cross-mamba mechanism to make the compensation stage more dynamic and flexible, preventing feature distortions.
Finally, we minimize the frequency domain gaps between reconstructed and ground-truth frames by proposing a simple yet effective Frequency Charbonnier-like loss 
that better preserves high-frequency content and enhances visual quality. Through extensive experiments, VSRM achieves state-of-the-art results on diverse benchmarks, establishing itself as a solid foundation for future research.
\end{abstract}    
\section{Introduction}
\label{sec:intro}

Video super-resolution (VSR) aims to produce higher resolution from low-resolution videos.
VSR typically processes long sequences and benefits from large receptive fields, efficiently capturing inter-frames information \cite{fuoli2023fast, chan2021basicvsr}. However, it incurs high computational costs, requiring a balance between performance and efficiency.
VSR usually uses convolutional neural network (CNN) \cite{chan2021basicvsr, jiang2020hierarchical, yang2021real} or transformer \cite{shi2022rethinking, zhou2024video, liang2024vrt, xu2024enhancing}. Tab. \ref{tab1} compares the pros and cons of these methods.
CNN-based methods cannot adjust based on input data and limit receptive fields \cite{dai2017deformable, su2019pixel} to capture inter-frame content.
Transformer-based methods show strong performance due to the power of the attention mechanism. However, full attention has quadratic complexity \cite{lu2021soft, hua2022transformer}, which is impractical for long sequences, while local attention struggles with limited receptive fields \cite{dong2022cswin}.
Recently, Gu \textit{et al.} introduced Mamba \cite{gu2023mamba}, which 
can model long-range information and data-dependent characteristics with linear complexity for NLP tasks. More recently, Mamba has been successfully applied in various vision tasks \cite{nasiri2024vim4path, guo2024mambair, yang2024vivim, chen2024video}. As far as we know, \textit{no research has employed Mamba to address VSR.} In this work, we propose VSRM, a novel Mamba-based framework for VSR. Specifically, we propose Dual Aggregation Mamba Block (DAMB), which contains $N$ Spatial-to-Temporal Mamba blocks (S2TMBs) followed by a Temporal-to-Spatial Mamba block (T2SMB) to model long-range spatio-temporal correlations for long sequences, enlarging large receptive fields. In S2TMB and T2SMB, we employ spatial-to-temporal and temporal-to-spatial scanning mechanisms to capture both spatial and temporal contexts.

The key challenge of VSR is effectively using the complementary information from adjacent frames, which may be misaligned due to object motions. To establish inter-frame correlations, many works employ the alignment module for VSR \cite{shi2022rethinking, xu2024enhancing, chan2022basicvsr++, chan2021basicvsr}. 
However, most studies use bilinear or nearest-neighbor interpolation for spatial alignment due to its simplicity.
Theoretically, using fixed weights (e.g., linear interpolation) causes feature distortions, leading to suboptimal results \cite{shi2022rethinking, chan2021understanding}. To address this, Xu \textit{et al.} \cite{xu2024enhancing}
proposed an attention-based implicit interpolation for the alignment module.
However, they computed interpolation in the fixed reference window, limiting flexibility and inaccuracy in aligning adjacent frames, especially in large motions between frames.
Based on this observation, we introduce Deformable Cross-Mamba Alignment (DCA), which dynamically aligns neighboring frames using a deformable cross-mamba mechanism that allows information exchange between multiple selective scan modules.


Current VSR methods use pixel-based losses (e.g., L1/L2 loss \cite{kim2016accurate, lim2017enhanced}, charbonnier loss \cite{chan2022basicvsr++, shi2022rethinking}) or perception-based losses (e.g., perceptual loss \cite{johnson2016perceptual, rad2019srobb}, adversarial loss \cite{ledig2017photo, park2018srfeat}).
Pixel-wise losses lead to overly smooth results due to their restriction in capturing perceptual similarity. Perception-based losses enhance visual quality by recovering finer details but introduce greater distortion. As VSR is an ill-posed task, gaps between super-resolution and ground-truth frames are inevitable, sometimes visible only in the frequency domain \cite{cai2021freqnet, hao2024learning}. Therefore, combining loss in the spatial and frequency domains may enhance the model's performance. We propose Frequency Charbonnier-like Loss (FCL), calculated from the real and imaginary parts of frequency components in Fourier space. FCL helps our model focus more on preserving and generating high-frequency details while maintaining low-frequency content.


\noindent
\textbf{Contribution.} \textbf{1)} We successfully adapt Mamba-based model for VSR for the first time and propose a new framework, VSRM.
VSRM introduces Dual Aggregation Mamba Block (DAMB), which aggregates spatial and
temporal features from sequences by using multi-scan direction mechanisms to obtain powerful representation ability and enlarge the model's receptive field.
\textbf{2)} We design Deformable Cross-mamba Alignment (DCA) module to provide a more flexible 
and learnable alignment through a deformable window cross-mamba mechanism in the compensation stage. \textbf{3)} We propose Frequency Charbonnier-like loss (FCL), which is computed in the frequency domain to preserve and generate high-frequency details better. \textbf{4)} VSRM outperforms all state-of-the-art (SOTA) methods on various benchmarks and shows the potential as the solid backbone for VSR.
\begin{table}[t]
\scalebox{0.78}{
\begin{tabular}{lcccc}
\toprule
\textbf{Model} & \textbf{\begin{tabular}[c]{@{}c@{}}Linear \\ Complexity\end{tabular}} & \textbf{\begin{tabular}[c]{@{}c@{}}Receptive \\ Field\end{tabular}} & \textbf{\begin{tabular}[c]{@{}c@{}}Data-\\ driven\end{tabular}} & \textbf{\begin{tabular}[c]{@{}c@{}}Notable \\ Method\end{tabular}} \\
\midrule
CNN              & \textcolor{green}{\faSmileO}                 & \textcolor{red}{\faFrownO}                     & \textcolor{red}{\faFrownO}           & BasicVSR \cite{chan2021basicvsr}              \\
Local Attention & \textcolor{green}{\faSmileO}                 & \textcolor{red}{\faFrownO}                      & \textcolor{green}{\faSmileO}           & IART \cite{xu2024enhancing}              \\
Full Attention   & \textcolor{red}{\faFrownO}                 & \textcolor{green}{\faSmileO}                      & \textcolor{green}{\faSmileO}           & -             \\
Mamba            & \textcolor{green}{\faSmileO}                 & \textcolor{green}{\faSmileO}                      & \textcolor{green}{\faSmileO}           & VSRM (ours)              \\ \bottomrule
\end{tabular}
}
\vspace{-2mm}
\caption{Comparison of the backbone for VSR. VSRM reaches both linear complexity and a large receptive field.}
\label{tab1}
\end{table}

\section{Related Work}
\label{sec:related_work}

\noindent
\textbf{Video Super-Resolution.} 
\textbf{1)} CNN-based methods \cite{kappeler2016video, tian2020tdan, chan2021basicvsr, yang2021real, luo2020video} is the early approach in VSR.
Initially, VSRnet \cite{kappeler2016video} extended CNN-based networks to VSR. Following this, TDAN \cite{tian2020tdan} introduced deformable convolution alignment to enhance the alignment module's performance. In contrast, BasicVSR \cite{chan2021basicvsr} proposed an effective model consisting solely of generic components, residual blocks, and optical flow, which surpasses all previous methods. 
\textbf{2)} Transformer-based methods \cite{cao2021video, liang2024vrt, shi2022rethinking, xu2024enhancing, zhou2024video, tran2024channel, tran2022trans2unet} achieve superior performance in VSR. Cao \textit{et al.} \cite{cao2021video} 
proposed spatial-temporal convolutional self-attention to explore better locality features. VRT \cite{liang2024vrt} used a temporal mutual self-attention for a wider temporal receptive field and a parallel warp module to handle large motion.
PSRT \cite{shi2022rethinking}  
introduced patch alignment that aligns image patches instead of pixels to maintain intra-patch relationships and preserve sub-pixel information. Based on this, Xu \textit{et al.} \cite{xu2024enhancing} proposed an implicit alignment module based on cross-attention mechanism to adjust feature representations across frames, further improving performance. These methods applied window-based attention with small window to avoid the quadratic complexity of full attention but were limited by local receptive fields, affecting cross-frame information capture. 
Recent methods, like PSRT \cite{shi2022rethinking}  and IART \cite{xu2024enhancing}, focused on designing window-based alignment modules rather than improving the transformer backbone to enhance performance. They employed a fixed window for implicit motion compensation, limiting the flexibility of alignment modules, especially when minimal or excessive motion. To address these issues, this study proposes a new framework, VSRM, based on Mamba to leverage its advantage in large receptive fields and linear complexity. We also design Deformable Cross-mamba Alignment for more flexible motion compensation using a deformable mechanism to boost performance.

\noindent
\textbf{State Space Model.} SSMs \cite{gu2021efficiently, smith2023simplified, gu2023mamba, fu2023hungry} 
show potential to replace Transformers \cite{tran2024channel, tran2022trans2unet}. Gu \textit{et al.} introduced S4 \cite{gu2021efficiently} that achieves linear complexity modeling of long sequences, followed by S5\cite{smith2023simplified} with improved capacity and efficiency.
Mamba \cite{gu2023mamba}, built upon SSMs, has recently surpassed Transformers on many benchmarks in the NLP domain due to its data-dependent characteristics and global receptive fields. In computer vision, Vim \cite{vim2024} adapted Mamba for images by converting them into sequences and using a bidirectional mamba block for global modeling. VideoMamba \cite{li2024videomamba} processed video by rearranging frames into spatio-temporal sequences, achieving comparable results to Transformer-based methods. MambaIR \cite{guo2025mambair} adopted Mamba for image restoration tasks, demonstrating that Mamba can work well for low-level vision tasks, such as video deblurring, denoising, etc. Expanding on that progress, \textit{we are pioneers in investigating the capabilities of Mamba for VSR}, utilizing its large receptive fields and linear complexity.

\begin{figure*}[t]
\begin{center}
\scalebox{0.80}{
\includegraphics[width=1.03\textwidth]{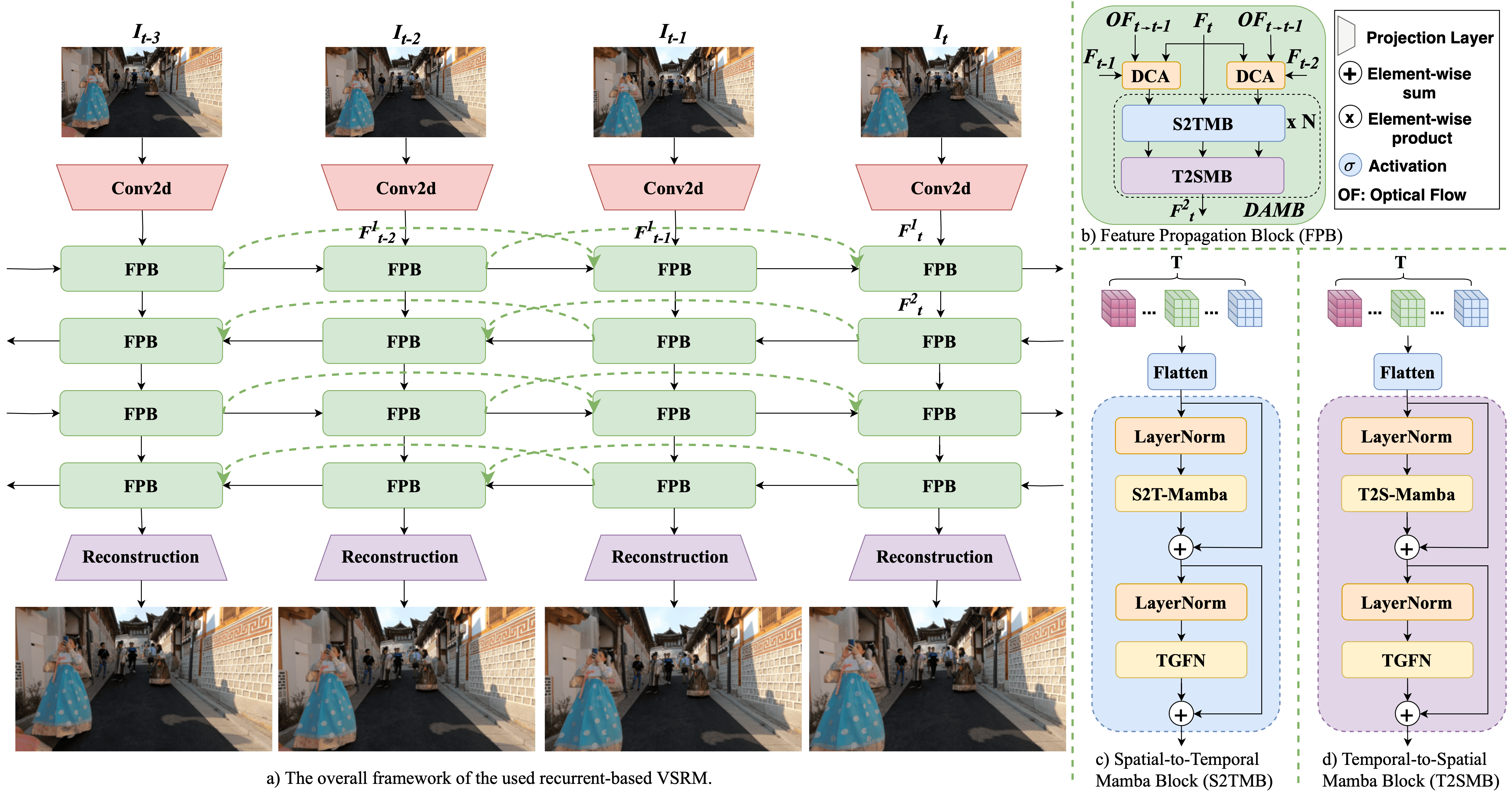}}
\end{center}
\vspace{-5mm}
   \caption{The network architecture of our method. a) Video Super-Resolution based on Mamba (VSRM). b) Feature Propagation block (FPB) consists of Deformable Cross-mamba Alignment (DCA) and Dual Aggregation Mamba Blocks (DAMB). c) Spatial-to-Temportal Mamba block (S2TMB). d) Temporal-to-Spatial Mamba block (T2SMB).}
\label{fig:arch}
\end{figure*}

\noindent
\textbf{Spectral bias.} Many studies reveal \textit{spectral bias} \cite{tancik2020fourier, rahaman2019spectral, mildenhall2021nerf} that models bias toward low-frequency rather than high-frequency content. Meanwhile, high-frequency content (e.g., edges, textures) is vital in low-level vision tasks \cite{jiang2021focal, fuoli2021fourier, kim2023whfl, korkmaz2024training}. Hence, Jiang \textit{et al.} proposed Focal Frequency Loss (FFL) \cite{jiang2021focal} computed in the frequency domain to emphasize high-frequency content and enhance visual quality. Based on this, Fuoli \textit{et al.} \cite{fuoli2021fourier} employed a GAN loss in Fourier space.
In contrast, WHFL \cite{kim2023whfl} used Wavelet transform, incorporating a module for multi-bandwidth analysis. These methods employ a frequency-adjusted reweight matrix 
to optimize the low and high-frequency elements.
However, the reweight matrix does not effectively balance the gradients of low and high frequencies. Furthermore, images mainly contain low-frequency content, so overemphasis can harm performance.
Hence, 
we propose Frequency Charbonnier-like loss (FCL), which directly computes loss on the \textit{Real} and \textit{Imaginary} parts in Fourier space. Unlike FFL \cite{jiang2021focal}, FCL treats low- and high-frequency components equally. 
\section{Preliminaries: State Space Models}
SSMs draw their inspiration from continuous linear time-invariant systems, which map a 1D function or sequence $x(t) \in \mathbb{R} \rightarrow y(t) \in \mathbb{R}$ via a hidden state $h(t) \in \mathbb{R}^{N}$. This is formulated as ordinary differential equations (ODE):
\begin{equation} \label{e:equation1}
h'(t) = Ah(t) + Bx(t), \;
y(t) = Ch(t), 
\end{equation}
where $A \in \mathbb{R}^{N\times N}$ is the evolution matrix, while $B \in \mathbb{R}^{N\times 1}$ and $C \in \mathbb{R}^{1\times N}$ are the projection parameters. To model discrete sequences like images, we need to approximate them via discretization. SSMs utilize the zero-order hold method, incorporating a timescale parameter $\Delta$ to map the continuous parameters $A$, $B$ to discrete parameters $\bar{A}$, $\bar{B}$. The process of discretization is outlined as follows:
\begin{subequations}
\begin{align}
\begin{split}
h_{t} = \bar{A}h_{t-1} + \bar{B}x_{t}, \; 
y_{t} = Ch_{t}, \label{e:equation2a}
\end{split}\\
\begin{split}
\bar{A} = exp(\Delta A), \;
\bar{B} = (\Delta A)^{-1}(exp(\Delta A - I))\Delta B, \label{e:equation2b}
\end{split}
\end{align}
\end{subequations}

The parameters of these SSMs are independent of the data, which means that $A$, $B$, and $C$ remain constant regardless of the input, thereby restricting their adaptability in sequence modeling. Recently, Mamba \cite{gu2023mamba} proposes selective SSMs (S6) with input data-dependent $x_{i} \in \mathbb{R}^{L}$:
\begin{equation} \label{e:equation3}
B_{i} = S_{B}x_{i}, \;
C_{i} = S_{c}x_{i}, \;
\Delta _{i} = Softplus(S_{\Delta}x_{i}), 
\end{equation}
where $S_{B}, S_{C} \in \mathbb{R}^{N\times L}$ and $S_{\Delta} \in \mathbb{R}^{L\times L}$ are linear projection layers. $B_{i}$ and $C_{i}$ are utilized across for all channels of $x_{i}$, $\Delta x_{i}$ contains $\Delta$ of $L$ channels, and $A$ remains consistent with earlier SSMs. Via the discretization presented in equations \ref{e:equation3}, $\bar{A}$ and $\bar{B}$ become input data-dependent. 
\section{Methodology}
In this section, we introduce VSRM architecture and its core components: Dual Aggregation Mamba block (DAMB) and Deformable Cross-mamba Alignment (DCA), as shown in Fig. 1. Furthermore, we present the training objectives to optimize our model.

\subsection{Overview}
 Let \( I_{t}^{LR} \in \mathbb{R}^{T\times H\times W\times C} \) 
 be the t-th low-resolution (LR) video frame, while \( I_{t}^{HR} \in \mathbb{R}^{T\times sH\times sW\times C} \) 
 denotes the t-th corresponding ground-truth (GT or HR) frame. Here, \( T, H, W, C, \) and \( s \) 
 indicate the total number of frames, height, width, channels, and the upscaling factor, respectively. Our aim is to generate the super-resolution video
 frame \( I_{t}^{SR} \in \mathbb{R}^{T\times sH\times sW\times C} \) 
 using the consecutive \( 2N+1 \) frames \( \{I^{t+N}_{i=t-N}\} \), 
 ensuring that \( I_{t}^{SR} \) and \( I_{t}^{HR} \) are as closely as possible.
 VSRM is depicted in Fig. \ref{fig:arch}a) includes two parts: Feature Extraction (Conv2d and Feature Propagation Block) and Upsampler (Reconstruction). Feature extraction extracts deep features from input frames and aligns and fuses the features of adjacent frames to leverage inter-frame relationships. The upsampler upscales the fused spatio-temporal features to produce the output sequence frames.

\subsection{Dual Aggregation Mamba Block}
\begin{figure}[t]
\begin{center}
\scalebox{0.90}{
\includegraphics[width=0.5\textwidth]{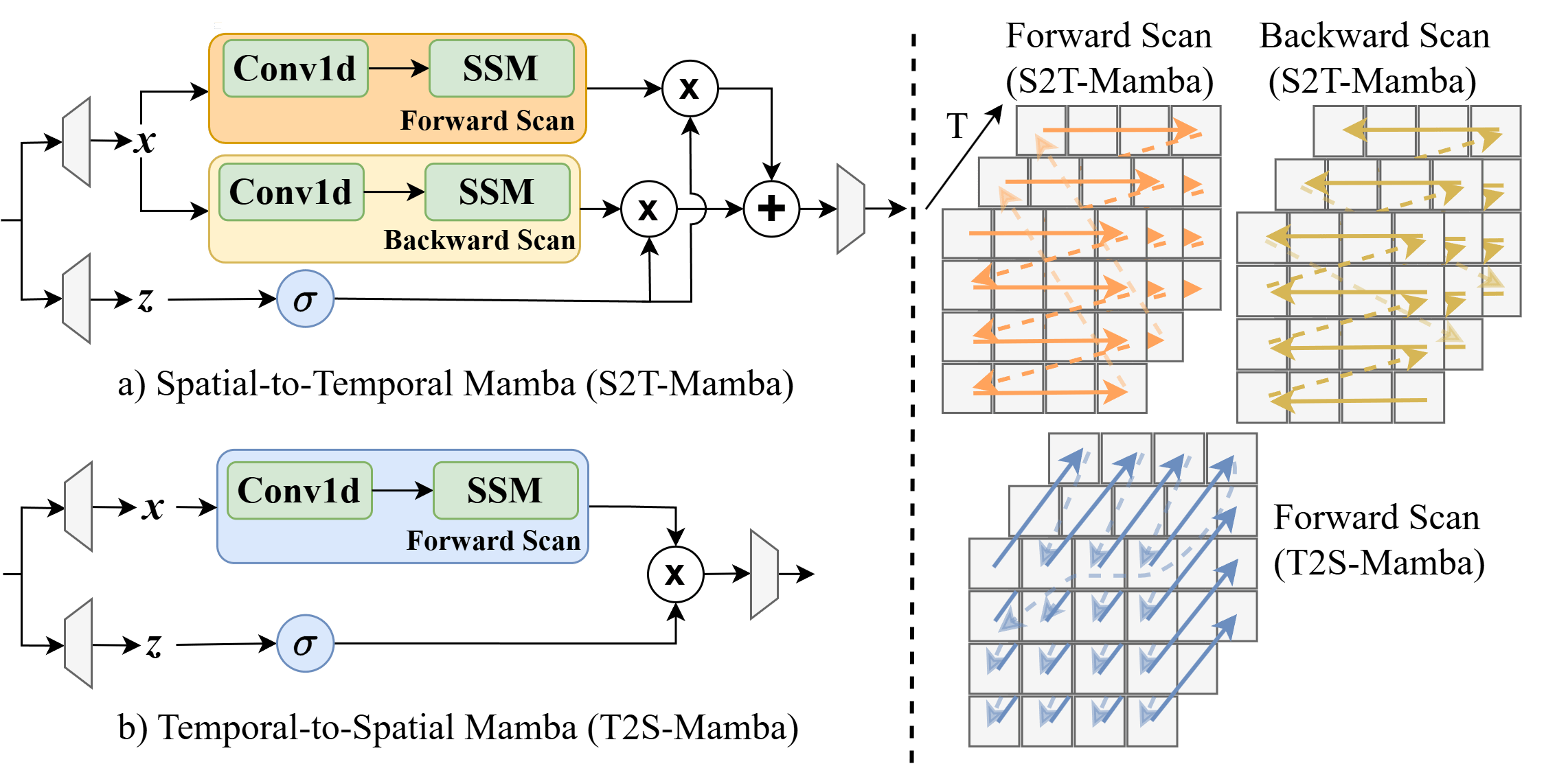}}
\end{center}
\vspace{-5mm}
   \caption{Structure and scan directions of Mamba module. a) Spatial-to-Temporal Mamba (S2T-Mamba). b) Temporal-to-Spatial Mamba (T2S-Mamba). For simplicity, we exclude the initial normalization and the final residual.}
\label{fig:direction_scan}
\end{figure}

Dual Aggregation Mamba block (DAMB) is the core component of VSRM that consists of Spatial-to-Temporal Mamba block (S2TMB) and Temporal-to-Spatial Mamba block (T2SMB), as shown in  Fig. \ref{fig:arch}(c, d). S2TMB and T2SMB are based on two modules: Spatial-to-Temporal Mamba (S2T-Mamba)  and Temporal-to-Spatial Mamba (T2S-Mamba), with their structures shown in Fig. \ref{fig:direction_scan}. By organizing $N$ S2TMBs followed by a T2SMB in DAMB, VSRM fully extracts spatio-temporal information in both spatial and temporal dimensions. Moreover, we propose Temporal Gated Feed-forward Network (TGFN) to explore spatial and temporal information better.

\noindent
\textbf{Spatial-to-Temporal Mamba.} 
S2T-Mamba structure and scan directions are shown in Fig. \ref{fig:direction_scan}(a). A 3D sequence 
($T$ frames) is converted into a 1D sequence and processed in two directions: \textcolor{BurntOrange}{forward} ($ForwardSSM(.)$)
and \textcolor{Goldenrod}{backward} ($BackwardSSM(.)$) paths.
This approach follows a mechanism that first scans spatially and then temporally, moving from frame 1 to frame $T$. This enables both spatial and temporal awareness during processing. The original Mamba was designed for 1D sequences, so this bidirectional scan mechanism helps preserve spatial awareness. By employing the scanning process from spatial to temporal in both the forward and backward paths, our module is optimized for handling 3D sequences more effectively. Given an input $X_{in}$, the output of S2T-Mamba is computed as:
\begin{subequations}
\begin{align}
\begin{split}
x = Linear(X_{in}), \;
z = \sigma (Linear(X_{in})), \label{e:s2t_1}
\end{split}\\
\begin{split}
x_1 = ForwardSSM(Conv1d(x)), \label{e:s2t_2}
\end{split}\\
\begin{split}
x_2 = BackwardSSM(Conv1d(x)), \label{e:s2t_3}
\end{split}\\
\begin{split}
S2T{\text -}Mamba(x, z) = Linear(x_1 \odot z + x_2 \odot z) \label{e:s2t_4}
\end{split}
\end{align}
\end{subequations}

\noindent
\textbf{Temporal-to-Spatial Mamba.} 
T2S-Mamba structure and scan direction are shown in Fig. \ref{fig:direction_scan}(b). T2S-Mamba also converts 3D sequence into a 1D sequence. However, we only perform the scanning in \textcolor{Blue}{forward} direction 
for two reasons. First, scanning in both directions could lead to redundant computations. Second, our experiment shows that using only forward scan in T2S-Mamba yields better results than adding backward pass, shown in Tab. \ref{tab:ab3}. We found that S2T-Mamba prioritizes extracting spatial rather than temporal information, leading to not extracting fully temporal information.
In contrast, T2S-Mamba explicitly prioritizes scanning temporal rather than spatial information, allowing for a more comprehensive feature extraction. 
Given an input $X_{in}$, the output of S2T-Mamba is calculated as:
\begin{subequations}
\begin{align}
\begin{split}
x = Linear(X_{in}), \;
z = \sigma (Linear(X_{in})), \label{e:t2s_1}
\end{split}\\
\begin{split}
x_1 = ForwardSSM(Conv1d(x)), \label{e:t2s_2}
\end{split}\\
\begin{split}
T2S{\text -}Mamba(x, z) = Linear(x_1 \odot z) \label{e:t2s_3}
\end{split}
\end{align}
\end{subequations}

\begin{figure}[t]
\begin{center}
\scalebox{0.65}{
\includegraphics[width=0.52\textwidth]{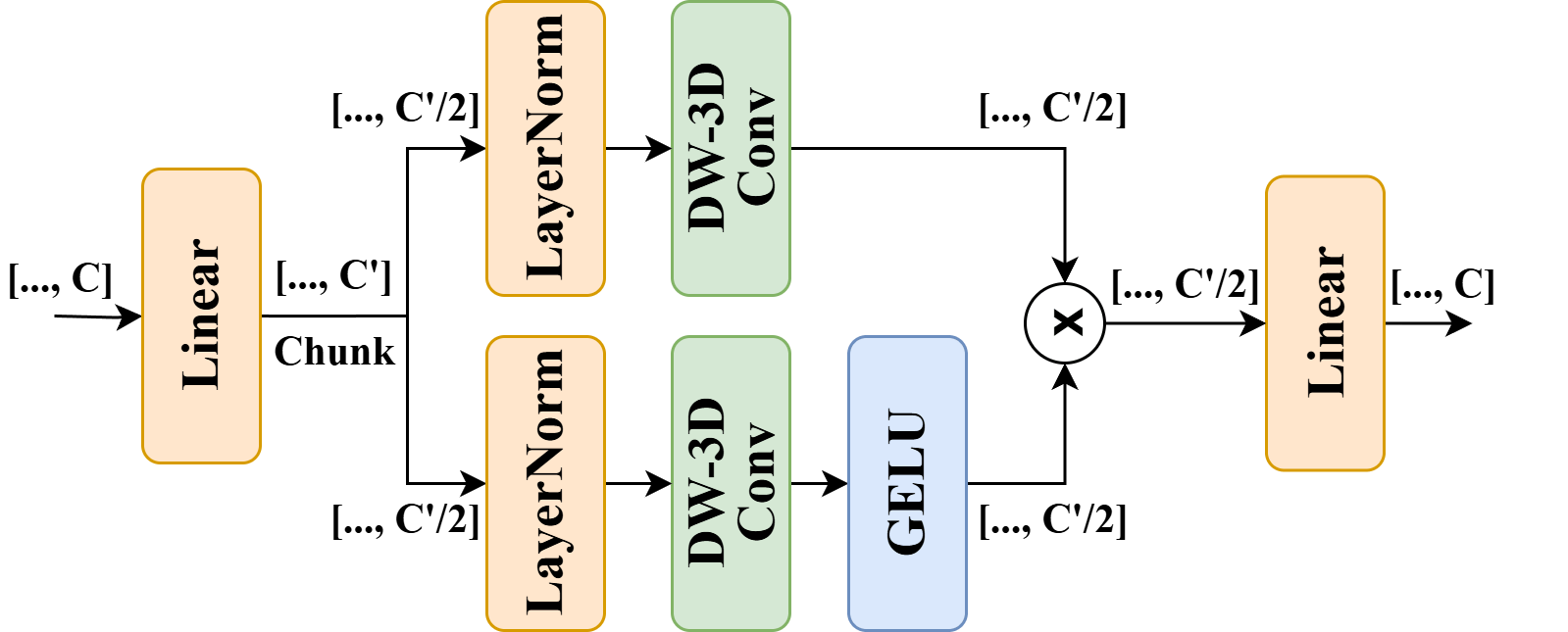}}
\end{center}
\vspace{-5mm}
   \caption{Temporal Gated Feed-forward Network (TGFN)}
\label{fig:ffn}
\end{figure}

\noindent
\textbf{Temporal Gated Feed-forward Network.} Feed-forward network (FFN) \cite{vaswani2017attention} consists of a nonlinear activation positioned between two linear projection layers. It disregards spatial and temporal relationships and lacks feature alignment
since it processes each input token individually. Moreover, the redundant information across channels hinders progress in learning feature representations.
To address this issues, we propose Temporal Gated Feed-forward Network (TGFN) shown in Fig. \ref{fig:ffn}. It has two main updates to FFN: 1) 3D depthwise convolutions (DW-3D Conv) and 2) a gating mechanism. We use DW-3D Conv to model relationships effectively from spatially and temporally neighboring pixel positions, which is vital in VSR.
TGFN employs a gating mechanism by performing element-wise multiplication on two parallel feature branches obtained from the chunking method, with one of them passed by the GELU, which has become the preferred option in recent work \cite{liang2021swinir, chen2023dual}.
Given the input $X \in \mathbb{R}^{T\times H\times W\times C}$, TGFN is formulated as:
\begin{subequations}
\begin{align}
\begin{split}
\resizebox{0.26\textwidth}{!}{
$\hat{X} = W^{1}_{p}X,$ \; 
$[\hat{X}_{1}, \hat{X}_{2}] = \hat{X}$} \label{e:equation4a}
 \end{split}\\
\begin{split}
\resizebox{0.42\textwidth}{!}{
$TGFN(X) = W^{2}_{p}(W^{1}_{d}LN(\hat{X}_{1}) \odot \sigma (W^{2}_{d}LN(\hat{X}_{2})))$} \label{e:equation4b}
\end{split}
\end{align}
\end{subequations}
where $W^{1}_{p}$, $W^{2}_{p}$ are weights of linear projections; $W^{1}_{d}$, $W^{2}_{d}$ are weights of the 3D DWConv layers; $\sigma$ is GELU activation, $C$ is the hidden dimension, LN is LayerNorm. 
Overall, TGFN adds nonlinear spatial and temporal information more effectively and optimizes the information flow through our pipeline.

\subsection{Deformable Cross-mamba Alignment}
\begin{figure}[t]
\begin{center}
\scalebox{0.74}{
\includegraphics[width=0.62\textwidth]{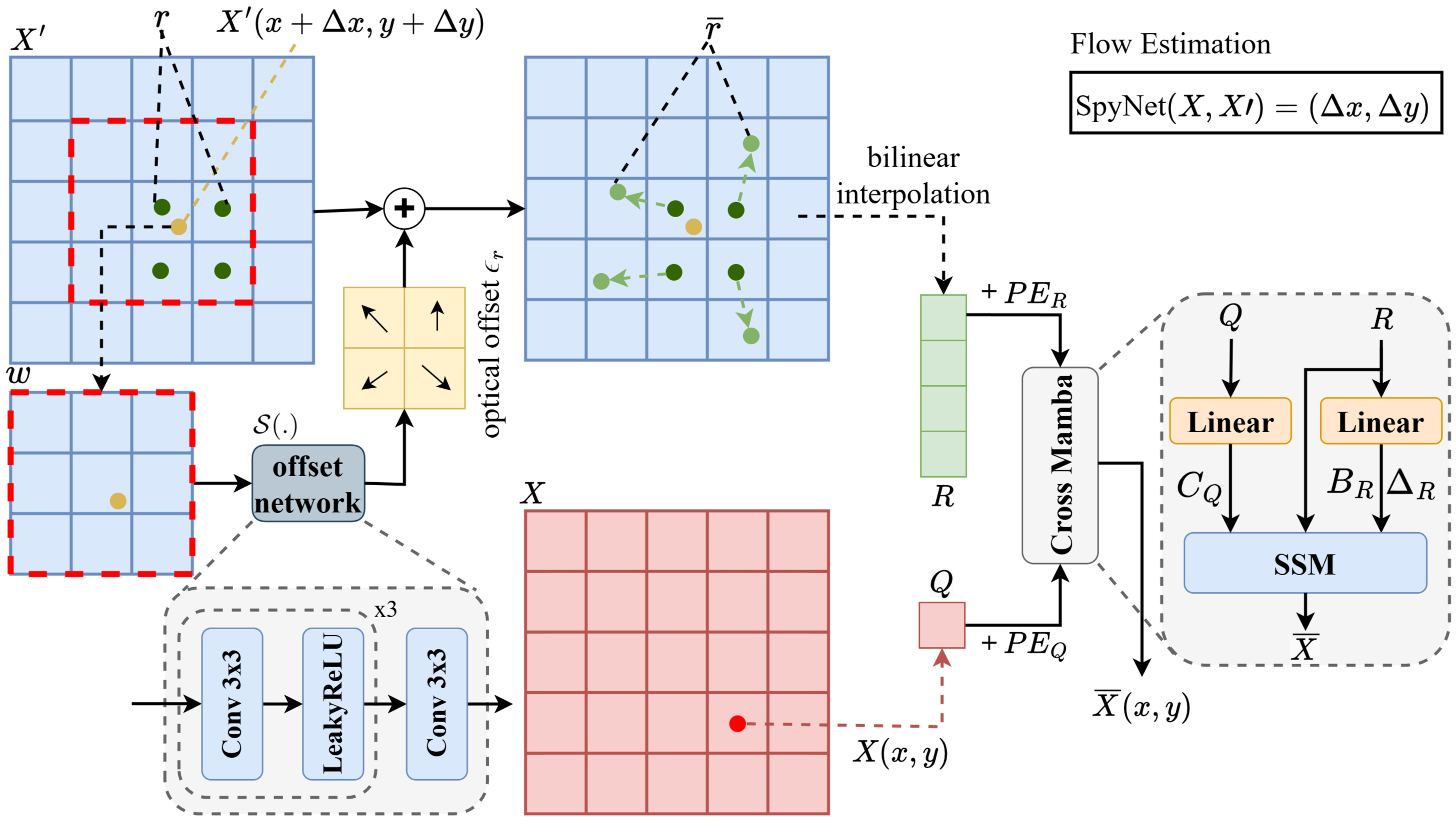}}
\end{center}
\vspace{-5mm}
   \caption{Pipeline of our proposed Deformable Cross-mamba Alignment (DCA). 
   We initialize a fixed reference region $r$ within the window $w$. This region is then adjusted to a dynamic reference region \(\bar r\) using an offset $\epsilon_r$, which is learned from the features within $w$.
   Finally, DCA learns the affinity through the cross-mamba module to calculate the final result (aligned point).}
\label{fig:aca}
\end{figure}

Alignment modules in VSR often fail 
when pixel movement between consecutive frames is minimal or excessive. 
This requires an alignment module adaptable to varying pixel movement. Building upon 
\cite{chan2021basicvsr, chan2022basicvsr++, shi2022rethinking}, 
we design Deformable Cross-mamba Alignment (DCA), as shown in Fig. \ref{fig:aca}. DCA uses the pre-trained SpyNet \cite{ranjan2016opticalflowestimationusing} for flow estimation while incorporating a new motion compensation scheme. Specifically, we use a deformable window scheme to perform a more flexible warping operation in the compensation stage between frames according to inter-frame motion information. We then employ a cross-mamba module, which enables interactions between the target point and the values of the dynamic reference points to infer the aligned point, leading to more adaptive and accurate alignment results.


For each pixel \(X(x,y)\) in the target frame \(X\) (\textcolor{red}{red} point), let $(\Delta_x,\Delta_y)$ be the optical flow estimated between \(X\) and the reference frame \(X'\).
The aligned frame \(\bar X\) is obtained by a spatial warping function \(\mathcal{W}\) with its parameters $\theta_{\mathcal{W}}$ as:
\begin{equation}
    \bar X(x, y) = \mathcal{W}(X',(x+\Delta x, y+\Delta y);\theta_{\mathcal{W}}),
\end{equation}
In the warping operation, we learn offset $\epsilon_r$ from window $w$ to identify dynamic reference region $\bar r$.
We extract \(w \in \mathbb{R}^{H_w \times W_w \times C}\), which is centered at the sampling position \((x', y')\) = $(x+\Delta x, y+\Delta y)$ in $X'$. Here, \(H_w\), \(W_w\), and \(C\) are the height, width, and number of channels of the window, respectively, as shown in Fig. \ref{fig:aca}.
Within this window, we initialize a reference region \(r\in \mathbb{R}^{H_r\times W_r\times2}\), by iterating over $i \in (0, 1..., H_r)$ and $j \in (0, 1,..., W_r)$, as follows:
\begin{equation}
\scalebox{0.98}{$
    r(i,j) = X'(\lfloor x' \rfloor -\lfloor \frac{W_r-1}{2} \rfloor + i, \lfloor y' \rfloor -\lfloor \frac{H_r-1}{2} \rfloor + j)
    $}
\end{equation}
To obtain the dynamic reference region \(\bar r\), we refine \(r\) using a small learnable offset network \(\mathcal{S}(.)\), which contains only convolution and activation layers (Fig. \ref{fig:aca}). This network learns the reference offset \(\epsilon_r\) from the features within the window \(w\). The final adaptive reference map is computed using bilinear interpolation \(\phi(.;.)\) as follows:
\begin{equation}
    \bar r = \phi(w;r+\epsilon_r), \quad \text{with } \epsilon_r = \mathcal{S}(w), 
\end{equation}
\noindent 
Finally, we apply implicit warping within the dynamic reference region to align each pixel by selectively cross-fusing reference point values with the target grid value at \(X(x,y)\) using the cross-mamba module \(cross\text{-}mamba(.,.)\):
\begin{equation}
    \bar X(x,y) = cross{\text -}mamba(R, Q),
\end{equation}

\noindent where $R$ is the reference tensor and $Q$ is the target tensor, both augmented with positional information. The reference tensor $R$ is computed by adding the flattened vector $\bar r$ to the positional encoding $PE_R$, and the target tensor $Q$ is obtained by adding the target value $X(x,y)$ to the positional embedding $PE_Q$. We use sinusoidal positional encoding to capture high-frequency content, which maps continuous input coordinates into a high-dimensional space \cite{mildenhall2021nerf} (more details in the \textit{supp.} file).
Based on Eq. \eqref{e:equation2a}, the cross-mamba module computes output \(\bar X\) at each step \(t\) as:
\begin{equation}
H_t = \bar{A}_{{R}}H_{t-1}+\bar{B}_{{R}}\bar{R}_t, \;
\bar X_t = C_{{Q}}H_t, \label{eq4.11}
\end{equation}
where \(\bar{A}_{R}\), \(\bar{B}_{R}\) are matrices dependent on \(R\), and $C_{{Q}}$ is a cross-modal matrix dependent on ${Q}$.

\begin{table*}[]
\centering
\begin{tabular}{l|c|c||cc|cc|cc}
\toprule
\multirow{2}{*}{\textit{Method}} &
  \multirow{2}{*}{\textit{Frames}} &
  \textit{\#Params} &
  \multicolumn{2}{c|}{\textit{REDS4}} &
  \multicolumn{2}{c|}{\textit{Vimeo-90K-T}} &
  \multicolumn{2}{c}{\textit{Vid4}} \\
 &
   &
  \textit{(M)} &
  \textit{PSNR} &
  \textit{SSIM} &
  \textit{PSNR} &
  \textit{SSIM} &
  \textit{PSNR} &
  \textit{SSIM} \\ \midrule
EDVR \cite{wang2019edvr} &
  5/7 &
  20.6 &
  \multicolumn{1}{c}{31.09} &
  0.8800 &
  \multicolumn{1}{c}{37.61} &
  0.9489 &
  \multicolumn{1}{c}{27.35} &
  0.8264 \\
VSR-T \cite{cao2021video} &
  5/7 &
  32.6 &
  \multicolumn{1}{c}{31.19} &
  0.8815 &
  \multicolumn{1}{c}{37.71} &
  0.9494 &
  \multicolumn{1}{c}{27.36} &
  0.8258 \\
VRT \cite{liang2024vrt} &
  6/- &
  30.7 &
  \multicolumn{1}{c}{31.60} &
  0.8888 &
  \multicolumn{1}{c}{-} &
  - &
  \multicolumn{1}{c}{-} &
  - \\
PSRT-recurrent \cite{shi2022rethinking} &
  6/- &
  10.8 &
  \multicolumn{1}{c}{31.88} &
  0.8964 &
  \multicolumn{1}{c}{-} &
  - &
  \multicolumn{1}{c}{-} &
  - \\
IART \cite{xu2024enhancing} &
  6/- &
  13.4 &
  \multicolumn{1}{c}{\underline{32.15}} &
  \underline{0.9010} &
  \multicolumn{1}{c}{-} &
  - &
  \multicolumn{1}{c}{-} &
  - \\
VSRM (ours) &
  6/- &
  17.1 &
  \multicolumn{1}{c}{\textbf{32.43}} &
  \textbf{0.9059} &
  \multicolumn{1}{c}{-} &
  - &
  \multicolumn{1}{c}{-} &
  - \\ \midrule
BasicVSR++ \cite{chan2022basicvsr++} &
  30/14 &
  7.3 &
  \multicolumn{1}{c}{32.39} &
  0.9069 &
  \multicolumn{1}{c}{37.79} &
  0.9500 &
  \multicolumn{1}{c}{27.79} &
  0.8400 \\
VRT \cite{liang2024vrt} &
  16/7 &
  35.6 &
  \multicolumn{1}{c}{32.19} &
  0.9006 &
  \multicolumn{1}{c}{38.20} &
  0.9530 &
  \multicolumn{1}{c}{27.93} &
  0.8425 \\
RVRT \cite{liang2022recurrent} &
  30/14 &
  10.8 &
  \multicolumn{1}{c}{32.75} &
  0.9113 &
  \multicolumn{1}{c}{38.15} &
  0.9527 &
  \multicolumn{1}{c}{27.99} &
  0.8462 \\
PSRT-recurrent \cite{shi2022rethinking} &
  16/14 &
  13.4 &
  \multicolumn{1}{c}{32.72} &
  0.9106 &
  \multicolumn{1}{c}{\underline{38.27}} &
  \underline{0.9536} &
  \multicolumn{1}{c}{28.07} &
  0.8485 \\
IART \cite{xu2024enhancing} &
  16/7 &
  13.4 &
  \multicolumn{1}{c}{\underline{32.90}} &
  \underline{0.9138} &
  \multicolumn{1}{c}{38.14} &
  0.9528 &
  \multicolumn{1}{c}{\underline{28.26}} &
  \underline{0.8517} \\
VSRM (ours) &
  16/7 &
  17.1 &
  \multicolumn{1}{c}{\textbf{33.11}} &
  \textbf{0.9162} &
  \multicolumn{1}{c}{\textbf{38.33}} &
  \textbf{0.9543} &
  \multicolumn{1}{c}{\textbf{28.44}} &
  \textbf{0.8552} \\ \bottomrule
\end{tabular}
\vspace{-2mm}
\caption{Quantitative comparison (PSNR$\uparrow$ and SSIM$\uparrow$) of our VSRM with other state-of-the-art methods at a scale of 4x. The best and second-best results are highlighted in \textbf{bold} and \underline{underlined}, and “-” indicates no test results.}
\label{tab:quantitative_result}
\end{table*}

\begin{figure*}[h]
\begin{center}
\scalebox{0.9}{
\includegraphics[width=0.93\textwidth, height=0.4\textheight]{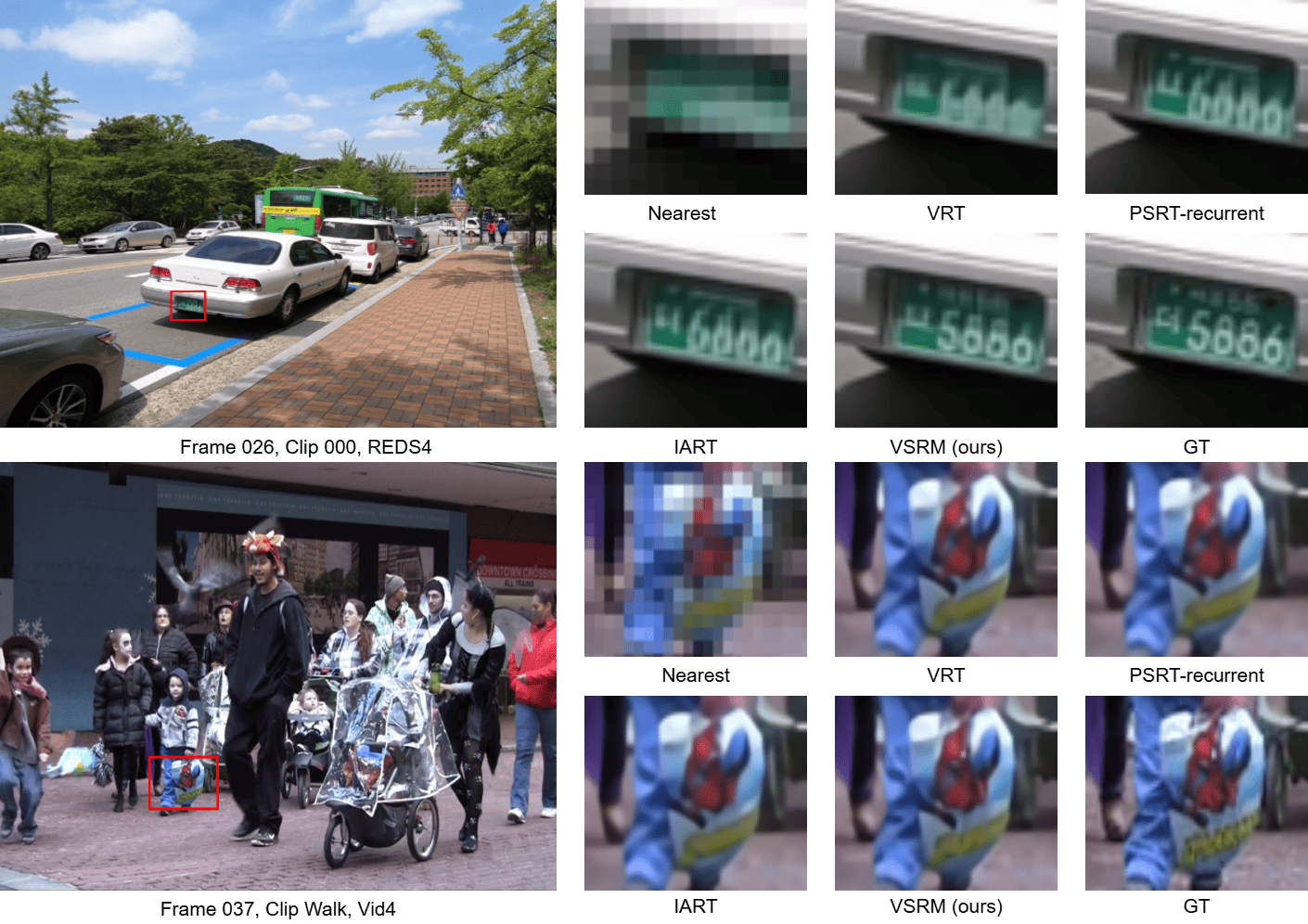}}
\end{center}
\vspace{-5mm}
\caption{Qualitative comparisons on REDS4 and Vid4 datasets, VSRM achieves clearer and more precise results, revealing finer patterns.}
\label{fig:viz}
\end{figure*}

\subsection{Training Objectives}
We use two loss functions, Charbonnier loss (CL) \cite{charbonnier1994two} and Frequency Charbonnier-like loss (FCL). CL minimizes the spatial domain's pixel gaps, while FCL reduces the frequency domain's gaps to recover high-frequency content. 

CL is considered a local loss, measuring the absolute difference between corresponding pixels of two frames without considering global structures. CL is as follows:
\begin{equation} \label{e:equation5}
\mathcal{L}_{CL} = \sqrt{\| \mathbf{I}_{SR} - \mathbf{I}_{HR} \|^2 + \epsilon^2},
\end{equation}

In contrast, FCL utilizes Fast Fourier Transformation (FFT) \cite{cooley1965algorithm} to convert frames from the spatial to the frequency domain. FFT utilizes the information from all pixels to obtain the FFT value for each spatial frequency. Thus, FCL is considered a global loss.
Given an image\footnotemark{} $I \in \mathbb{R}^{H\times W}$ of size $H\times W$, Discrete Fourier Transform $\mathcal{F}$ is as follows:
\footnotetext{For simplicity, we only present the formulation for gray images.}
\begin{equation} \label{e:equation6}
\resizebox{0.49\textwidth}{!}{
$
\mathcal{X}(u, v) = \mathcal{F}[\mathbf{I}] = \sum_{x=1}^{H} \sum_{y=1}^{W} \mathbf{I}(x, y) e^{-j 2 \pi \left( \frac{u}{H}x + \frac{v}{W}y \right)}
$
}
\end{equation}
$\mathbf{I}(x, y)$ is the pixel value at $(x, y)$; $\mathcal{X}(u, v)$ is complex value at frequency $(u, v)$; $e$ and $j$ are Euler’s number and the imaginary unit.
\noindent
$\mathcal{F}(.)$ is a complex number with \textit{Real} (\textit{Re}$\mathcal{F}(.)$) and \textit{Imaginary} (\textit{Im}$\mathcal{F}(.)$). We observe a limitation in calculating loss based on magnitude and phase. The magnitude calculation involves a square root, while the phase employs a nonlinear arctangent function. This creates discontinuities and increases sensitivity to numerical inaccuracies for VSR. To address this, we calculate frequency loss separately for the \textit{Real} and \textit{Imaginary}. This ensures that even minor changes in the spatial domain lead to proportional changes in these components, as they represent linear decompositions of the signal in Fourier space. The formulation of FCL and total loss are as follows:
\begin{equation}
\mathcal{L}_{FCL} = \! \! \! \! \sum_{i \in \{Re, Im\}} \! \! \! \! \! \! \lambda _{i} \sqrt{\| i\mathcal{F}(\mathbf{I}_{SR}) - i\mathcal{F}(\mathbf{I}_{HR}) \|^2 + \epsilon^2} 
\label{e:equation5}
\end{equation}
\begin{equation} \label{e:equation5}
\mathcal{L}_{total} = \lambda \mathcal{L}_{CL} + \mathcal{L}_{FCL},
\end{equation}
$\lambda, \lambda _{Re},$ and $\lambda _{Im}$ are hyperparameters; $\lambda _{Re}$, $\lambda _{Im} \ll \lambda$. $\lambda _{Re},$ and $\lambda _{Im}$ control the regularization strength of the high-frequency content in the output. 
In this work,  we set $\lambda$ = 1.0, $\lambda _{Re}$ = $\lambda _{Im}$ = 0.02 and $\epsilon$ = $10^{-3}$ in our experiments.
\section{Experiments}
\subsection{Datasets and Metrics}

\noindent
We use REDS \cite{nah2019ntire} and Vimeo-90K \cite{xue2019video} as training datasets. Following \cite{wang2019edvr}, for REDS, we use REDS4\footnotemark{} as a test set, and the remaining videos are used for training\footnotetext{Clips 000, 011, 015, 020 of REDS training set.}. For Vimeo-90K, Vid4 \cite{liu2013bayesian} and Vimeo-90K-T \cite{xue2019video} are used as test sets. These two training sets have different motion movements. The motion in Vimeo-90K is generally small, whereas REDS contains large motions with more pixel movement.
We study ×4 VSR task, use bicubic interpolation to produce LR video frames, and use PSNR/SSIM \cite{wang2004image} as the evaluation metrics.


\subsection{Comparisons with State-of-the-Art Methods}

We compare our VSRM with the SOTA VSR methods: EDVR \cite{wang2019edvr}, VSR-T \cite{cao2021video}, VRT \cite{liang2024vrt}, BasicVSR++ \cite{chan2022basicvsr++}, RVRT \cite{liang2022recurrent}, PSRT-recurrent \cite{shi2022rethinking}, IART \cite{xu2024enhancing}. 

\noindent
\textbf{Quantitative Results.} As shown in Tab. \ref{tab:quantitative_result}, VSRM achieves SOTA results across all benchmarks. Notably, VSRM surpasses VRT \cite{liang2024vrt}, a window-based transformer approach, by margins of up to 0.83dB and 0.90dB in PSNR for six and sixteen input frame settings on REDS4 while featuring a lower parameter count. Compared to the recent SOTA methods PSRT-recurrent\cite{shi2022rethinking} and IART \cite{xu2024enhancing}, VSRM demonstrates improvements of 0.4dB and 0.28dB in PSNR for six-frame input setting and 0.39dB and 0.21dB in PSNR for sixteen-frame input setting on REDS4, though it has slightly more parameters. 
These findings indicate that our method balances well between complexity and performance.  
Additionally, VSRM is effective for slow-motion datasets, including Vimeo-90K-T and Vid4, where our method also surpasses PSRT-recurrent \cite{shi2022rethinking} and IART \cite{xu2024enhancing} on these datasets for sixteen-frame input settings. This shows that our method works well in varying pixel movement and scenarios.
Overall, these gains represent a substantial improvement in VSR.

\begin{figure}[t]
\begin{flushleft}
\hspace{-3.5mm} 
\scalebox{0.80}{
\includegraphics[width=0.61\textwidth]{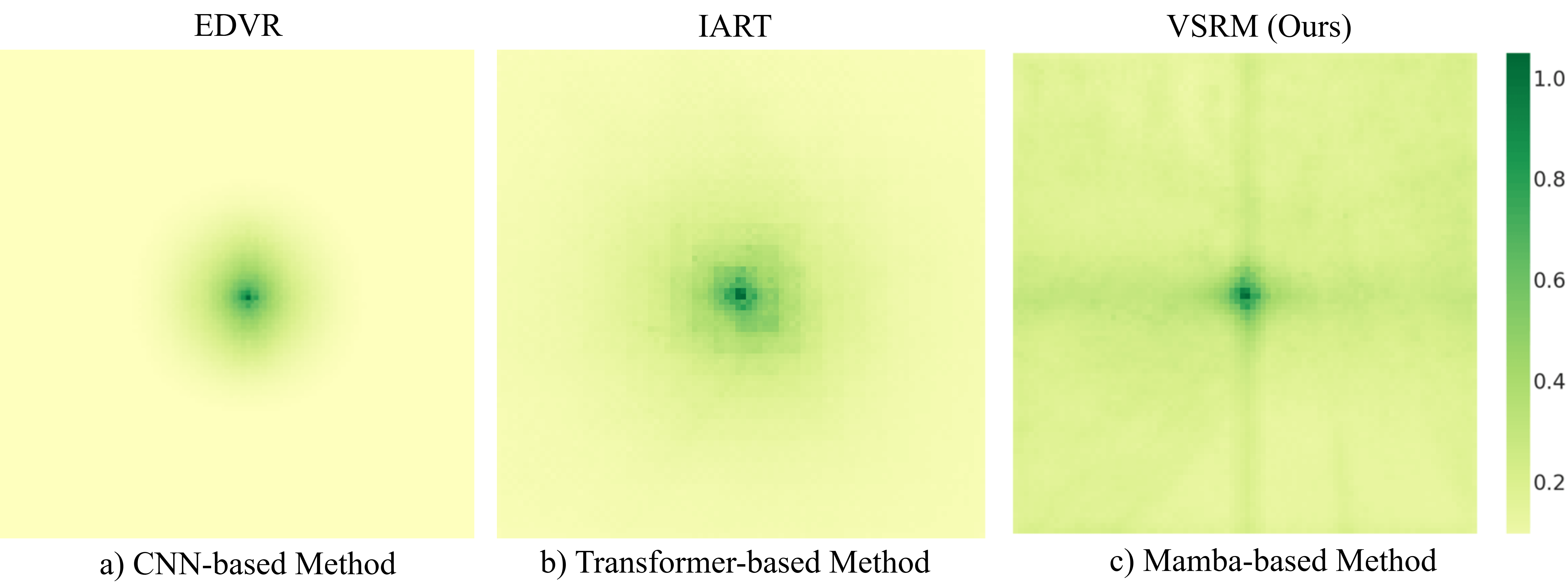}}
\end{flushleft}
\vspace{-5mm}
   \caption{Effective Receptive Field (ERF) \cite{luo2016understanding, ding2022scaling} visualization results. A more widely spread dark region signifies a larger receptive field. VSRM attains a considerable global receptive field.}
\label{fig:ERF}
\end{figure}

\noindent
\textbf{Visual Results.} Qualitative comparisons can be observed in Fig. \ref{fig:viz} and Fig. \ref{fig:ERF}. VSRM effectively captures intricate details. Specifically, VSRM is the only method that accurately restores the license plate numbers in the REDS4 dataset and the fine details of Spider-Man in the Vid4 dataset in Fig. \ref{fig:viz}. More examples of REDS4 and Vid4 datasets are shown in the \textit{supp.} file. Fig. \ref{fig:ERF} shows the effective receptive field of our method in comparison to EDVR \cite{wang2019edvr} (CNN-based) and IART \cite{xu2024enhancing} (Transformer-based) on REDS4.
A local receptive field restricts the CNN-based methods. In contrast, Transformer-based methods can enhance the receptive field but still face limitations due to the window attention mechanism for lower computational costs. Our method stands out as it successfully achieves a significantly larger global receptive field, enhancing our model's performance in VSR.

\subsection{Ablation Study}

We conduct ablation studies using VSRM with custom settings for efficiency to understand the contributions of each proposed module. Details settings are in the \textit{supp.} file.

\begin{table}[h]
\centering
\scalebox{0.88}{
\begin{tabular}{lccc}
\toprule
Model            & \multicolumn{1}{c}{PSNR (dB)} & Param. (M) & FLOPs (G) \\ \hline
3D DW-Conv       & 30.84                        & 19.49            & 149.8            \\
Window Attn. & 30.97                        & 7.68            & 152.4            \\
Full Attn   & 31.06                        & 7.68            & 1018.1            \\
Mamba (ours)            & \textbf{31.09}                        & 8.61            & 159.2            \\ \bottomrule 
\end{tabular}}
\vspace{-2mm}
\caption{Effective of Mamba modules.}
\label{tab:ab1}
\end{table}
\noindent
\textbf{Effect of Mamba modules.} Mamba modules are the core contribution of this study; they balance computational cost and high performance for inter-frame modeling. To verify its effectiveness, we replace these modules with convolutions from BasicVSR \cite{chan2021basicvsr}, window attention from IART \cite{xu2024enhancing}, and global attention as in Tab. \ref{tab:ab1}. 
Compared to Convolution and Window Attention methods, our method achieves notable performance gains, increasing PSNR by 0.25dB and 0.12dB, respectively, without adding significant computational overhead.
Our method reaches a similar PSNR with Full Attention but is much more efficient regarding model complexity. Overall, Mamba modules balance computational efficiency and performance well.

\begin{table}[h]
\centering
\scalebox{0.88}{
\begin{tabular}{lcccc}
\toprule
Structure & \multicolumn{1}{l}{PSNR (dB)}                  & Param. (M) & FLOPs (G) \\ \hline
w/o Align.             & 30.87                        & 8.53                        & 120.4          \\
FGDA \cite{liang2022recurrent}       & 30.92       & 8.70                     & 154.3        \\
IA \cite{xu2024enhancing}        & 31.00                & 8.57             & 148.7       \\ 
DCA (ours)        & \textbf{31.09}                       & 8.61                        & 159.2     \\
\bottomrule 
\end{tabular}}
\vspace{-2mm}
\caption{Effect of Deformable Cross-mamba Alignment module.}
\label{tab:ab2}
\end{table}

\noindent
\textbf{Effect of DCA.} Tab. \ref{tab:ab2} compares our DCA with existing alignment modules. Removing the alignment module significantly decreases performance ($\downarrow$ 0.22dB). 
Our DCA achieves higher PSNR than SOTA methods like FGDA \cite{liang2022recurrent} and IA \cite{xu2024enhancing} without incurring excessive computational overhead. Specifically, DCA achieves 0.17dB and 0.09dB higher PSNR than FGDA and IA, respectively, demonstrating the effectiveness of our proposed module. By leveraging the flexibility of the deformable window when computing cross-mamba, DCA enhances the precision of aligning adjacent frames to the target frame.


\begin{table}[h]
\centering
\scalebox{0.88}{
\begin{tabular}{lccc}
\toprule
Structure    & w/o T2SMB     & w/ T2SMB$^{\dag}$ & w/ T2SMB     \\ \hline
PSNR (dB)  & 30.95 & 31.02 & \textbf{31.09} \\
Param. (M) &       7.87           &           8.65  & 8.61      \\
FLOPs (G) &        155.6          &         162.2    &  159.2    \\ \bottomrule 
\end{tabular}}
\vspace{-2mm}
\caption{Effect of Temporal-to-Spatial Mamba block. $\dag$ indicates that both forward and backward scans are applied to T2SMB.}
\label{tab:ab3}
\end{table}

\noindent
\textbf{Effect of T2SMB.} Tab. \ref{tab:ab3} shows the effect of the Temporal-to-Spatial Mamba block (T2SMB). Removing T2SMB considerably reduces the model’s performance ($\downarrow$ 0.14dB), highlighting the block's importance. This performance drop occurs because the Spaital-to-Temporal Mamba block (S2TMB) cannot fully extract temporal information from the sequential input features. The T2SMB block prioritizes and explicitly focuses on extracting temporal information, enabling more comprehensive final features to be generated. We also conducted an ablation study with T2SMB using both forward and backward scans. Our findings indicate that using only the forward scan for T2SMB yields better performance while requiring less computational effort.

\begin{table}[h]
\centering
\scalebox{0.88}{
\begin{tabular}{lccc}
\toprule
Structure    & PSNR (dB)     & Param. (M) & FLOPs (G)      \\ \hline
FFN \cite{vaswani2017attention}     &     30.90     &   8.68   & 136.2        \\
TGFN w/o DWConv &       30.81       &        8.27      &    124.7       \\
TGFN w/o Split &        31.05          &         9.41    &    186.1     \\ 
TGFN (ours) &        \textbf{31.09}          &         8.61     &    159.2    \\ 
\bottomrule 
\end{tabular}}
\vspace{-2mm}
\caption{Effect of Temporal Gated Feed-forward Network.}
\label{tab:ab4}
\end{table}

\noindent
\textbf{Effect of TGFN.} Tab. \ref{tab:ab4} shows the effect of the temporal gated feed-forward network (TGFN). We compare models using regular FFN \cite{vaswani2017attention}, TGFN without 3D depthwise convolution (TGFN w/o DWConv), TGFN without split channel (TGFN w/o Split), and our TGFN. Compared to FFN, TGFN has a similar number of parameters and somewhat higher FLOPs while substantially improving performance ($\uparrow$ 0.19dB). Moreover, the performance suffers significantly ($\downarrow$ 0.28dB) when the 3D DWConv is removed from TGFN, emphasizing the necessity of both spatial and temporal information. Eliminating the split operation in TGFN leads to a slight decrease in PSNR but increases model size and complexity. This indicates that redundancy in channel features negatively impacts the models' performance.

\begin{table}[h]
\centering
\scalebox{0.88}{
\begin{tabular}{lcccccc}
\toprule
\multirow{2}{*}{Loss} & FFL                       & FSL\footnotemark{}                   & WHFL                  & \multicolumn{3}{c}{FCL (ours)} \\ \cline{5-7} 
                      & \cite{jiang2021focal}                         & \multicolumn{1}{c}{\cite{fuoli2021fourier}} & \multicolumn{1}{c}{\cite{kim2023whfl}} & 0      & 0.02   & 0.2   \\ \hline
PSNR                  & \multicolumn{1}{c}{31.01} & 30.99   &    30.90         & 30.97  & \textbf{31.09}  & 31.02 \\ \bottomrule 
\end{tabular}}
\vspace{-2mm}
\caption{Effect of Frequency Charbonnier-like loss (FCL). $i \in \{Re, Im\}$ and we set $\lambda _{Re} = \lambda _{Im}$.}
\label{tab:ab5}
\end{table}
\footnotetext{FSL: Fourier Space Loss}

\noindent
\textbf{Effect of FCL loss.} Tab. \ref{tab:ab5} shows the effectiveness of Frequency Chabornier-like loss (FCL). We compare our loss with FFL \cite{jiang2021focal}, FSL \cite{fuoli2021fourier}, and WHFL \cite{kim2023whfl} (default weight for these losses are 0.02 for fair comparison). Our FCL achieves a higher PSNR than FFL ($\uparrow$ 0.08dB), indicating that balancing low- and high-frequency components and using Real and Imaginary parts for loss calculation 
yields more optimal results. Moreover, our loss demonstrates higher performance compared to FSL and WHFL, with clear margins. These results show the effectiveness of our proposed FCL compared to other frequency losses.
Our method achieves the best results with 
 FCL ($\lambda _{i} = 0.02$ where $i \in \{Re, Im\}$). If FCL ($\lambda _{i} = 0$) is eliminated, performance substantially declines ($\downarrow$ 0.12dB), highlighting the importance of this loss.

\subsection{Model Complexity Comparison}

Tab. \ref{tab:ab6} reports comparing parameters, FLOPs, and runtime for VSRM and other VSR methods. The runtime is measured on a RTX A6000 48GB. Our VSRM is more efficient than EDVR when considering the number of parameters and FLOPs. Although it has a moderately higher complexity than Transformer-based methods that use window-based attention, these trade-offs are reasonable when balancing performance with model complexity. Furthermore, since the acceleration and optimization of Mamba are still being explored, our method has potential for further improvements.
\begin{table}[h]
\scalebox{0.88}{
\begin{tabular}{lccc}
\toprule
Method         & Param. (M)                & FLOPs (G) & Runtime (ms) \\ \hline
VSR-T \cite{cao2021video}           & \multicolumn{1}{c}{32.6}  & 177.2     & -        \\
EDVR \cite{wang2019edvr}           & 20.6                      & 323.6     & -        \\
PSRT-recurrent \cite{shi2022rethinking} & 13.4                      & 164.6     & 173        \\
IART \cite{xu2024enhancing}           & 13.4                     & 177.8     & 180        \\
VSRM (ours)     & \multicolumn{1}{c}{17.1} & 217.4    & 223 \\ \bottomrule 
\end{tabular}}
\vspace{-2mm}
\caption{Comparison of model complexity of different VSR models on the REDS4 dataset.}
\label{tab:ab6}
\end{table}
\vspace{-5mm}
\section{Conclusion}
In this paper, we introduce VSRM, which adapts Mamba for VSR for the first time and achieves state-of-the-art results on multiple benchmarks. Specifically, we propose Spatial-to-Temporal Mamba and Temporal-to-Spatial Mamba blocks to capture long-range spatio-temporal features effectively.
Next, we propose a novel Deformable Cross-Mamba Alignment that uses a deformable cross-mamba mechanism to adapt to varying pixel movement flex to enhance model performance compared to previous alignment modules. Finally, we introduce a Frequency Charbonnier-like loss to reduce the gaps between reconstructed and ground truth frames in the frequency domain and better recover high-frequency content. This demonstrates further improvement by utilizing both spatial and frequency losses. VSRM has the potential to establish a solid baseline for upcoming research and shows great promise in substituting Transformer-based models. Furthermore, VSRM could be adaptable for other low-level vision tasks, including video colorization, deblurring, and denoising.
\section*{Acknowledgments}
\label{sec:acknowledgments}
This work was supported by the Institute of Information \& Communications Technology Planning \& Evaluation(IITP)-Innovative Human Resource Development for Local Intellectualization program grant funded by the Korea government(MSIT)(IITP-2025-RS-2020-II201489). This work was also supported by the National Research Foundation of Korea(NRF) grant funded by the Korea government(MSIT)(RS-2025-00573160), and the Technology Innovation Program (RS-2025-02222776) funded By the Ministry of Trade, Industry \& Energy(MOTIE, Korea).

This work was also supported by Hyundai Motor Chung
Mong-Koo Global Scholarship to Dinh Phu Tran (first
author) and generous GPU support by Korea Association for ICT Promotion.
{
    \small
    \bibliographystyle{ieeenat_fullname}
    \bibliography{main}

\begin{thebibliography}{61}
\providecommand{\natexlab}[1]{#1}
\providecommand{\url}[1]{\texttt{#1}}
\expandafter\ifx\csname urlstyle\endcsname\relax
  \providecommand{\doi}[1]{doi: #1}\else
  \providecommand{\doi}{doi: \begingroup \urlstyle{rm}\Url}\fi

\bibitem[Cai et~al.(2021)Cai, Ding, and Lu]{cai2021freqnet}
Runyuan Cai, Yue Ding, and Hongtao Lu.
\newblock Freqnet: A frequency-domain image super-resolution network with dicrete cosine transform.
\newblock \emph{arXiv preprint arXiv:2111.10800}, 2021.

\bibitem[Cao et~al.(2021)Cao, Li, Zhang, and Van~Gool]{cao2021video}
Jiezhang Cao, Yawei Li, Kai Zhang, and Luc Van~Gool.
\newblock Video super-resolution transformer.
\newblock \emph{arXiv preprint arXiv:2106.06847}, 2021.

\bibitem[Chan et~al.(2021{\natexlab{a}})Chan, Wang, Yu, Dong, and Loy]{chan2021basicvsr}
Kelvin~CK Chan, Xintao Wang, Ke Yu, Chao Dong, and Chen~Change Loy.
\newblock Basicvsr: The search for essential components in video super-resolution and beyond.
\newblock In \emph{Proceedings of the IEEE/CVF conference on computer vision and pattern recognition}, pages 4947--4956, 2021{\natexlab{a}}.

\bibitem[Chan et~al.(2021{\natexlab{b}})Chan, Wang, Yu, Dong, and Loy]{chan2021understanding}
Kelvin~CK Chan, Xintao Wang, Ke Yu, Chao Dong, and Chen~Change Loy.
\newblock Understanding deformable alignment in video super-resolution.
\newblock In \emph{Proceedings of the AAAI conference on artificial intelligence}, pages 973--981, 2021{\natexlab{b}}.

\bibitem[Chan et~al.(2022)Chan, Zhou, Xu, and Loy]{chan2022basicvsr++}
Kelvin~CK Chan, Shangchen Zhou, Xiangyu Xu, and Chen~Change Loy.
\newblock Basicvsr++: Improving video super-resolution with enhanced propagation and alignment.
\newblock In \emph{Proceedings of the IEEE/CVF conference on computer vision and pattern recognition}, pages 5972--5981, 2022.

\bibitem[Charbonnier et~al.(1994)Charbonnier, Blanc-Feraud, Aubert, and Barlaud]{charbonnier1994two}
Pierre Charbonnier, Laure Blanc-Feraud, Gilles Aubert, and Michel Barlaud.
\newblock Two deterministic half-quadratic regularization algorithms for computed imaging.
\newblock In \emph{Proceedings of 1st international conference on image processing}, pages 168--172. IEEE, 1994.

\bibitem[Chen et~al.(2024)Chen, Huang, Xu, Pei, Chen, Li, Wang, Li, Lu, and Wang]{chen2024video}
Guo Chen, Yifei Huang, Jilan Xu, Baoqi Pei, Zhe Chen, Zhiqi Li, Jiahao Wang, Kunchang Li, Tong Lu, and Limin Wang.
\newblock Video mamba suite: State space model as a versatile alternative for video understanding.
\newblock \emph{arXiv preprint arXiv:2403.09626}, 2024.

\bibitem[Chen et~al.(2023)Chen, Zhang, Gu, Kong, Yang, and Yu]{chen2023dual}
Zheng Chen, Yulun Zhang, Jinjin Gu, Linghe Kong, Xiaokang Yang, and Fisher Yu.
\newblock Dual aggregation transformer for image super-resolution.
\newblock In \emph{Proceedings of the IEEE/CVF international conference on computer vision}, pages 12312--12321, 2023.

\bibitem[Cooley and Tukey(1965)]{cooley1965algorithm}
James~W Cooley and John~W Tukey.
\newblock An algorithm for the machine calculation of complex fourier series.
\newblock \emph{Mathematics of computation}, 19\penalty0 (90):\penalty0 297--301, 1965.

\bibitem[Dai et~al.(2017)Dai, Qi, Xiong, Li, Zhang, Hu, and Wei]{dai2017deformable}
Jifeng Dai, Haozhi Qi, Yuwen Xiong, Yi Li, Guodong Zhang, Han Hu, and Yichen Wei.
\newblock Deformable convolutional networks.
\newblock In \emph{Proceedings of the IEEE international conference on computer vision}, pages 764--773, 2017.

\bibitem[Ding et~al.(2022)Ding, Zhang, Han, and Ding]{ding2022scaling}
Xiaohan Ding, Xiangyu Zhang, Jungong Han, and Guiguang Ding.
\newblock Scaling up your kernels to 31x31: Revisiting large kernel design in cnns.
\newblock In \emph{Proceedings of the IEEE/CVF conference on computer vision and pattern recognition}, pages 11963--11975, 2022.

\bibitem[Dong et~al.(2022)Dong, Bao, Chen, Zhang, Yu, Yuan, Chen, and Guo]{dong2022cswin}
Xiaoyi Dong, Jianmin Bao, Dongdong Chen, Weiming Zhang, Nenghai Yu, Lu Yuan, Dong Chen, and Baining Guo.
\newblock Cswin transformer: A general vision transformer backbone with cross-shaped windows.
\newblock In \emph{Proceedings of the IEEE/CVF conference on computer vision and pattern recognition}, pages 12124--12134, 2022.

\bibitem[Fu et~al.(2023)Fu, Dao, Saab, Thomas, Rudra, and R{\'e}]{fu2023hungry}
Daniel~Y. Fu, Tri Dao, Khaled~K. Saab, Armin~W. Thomas, Atri Rudra, and Christopher R{\'e}.
\newblock Hungry {H}ungry {H}ippos: Towards language modeling with state space models.
\newblock In \emph{International Conference on Learning Representations}, 2023.

\bibitem[Fuoli et~al.(2021)Fuoli, Van~Gool, and Timofte]{fuoli2021fourier}
Dario Fuoli, Luc Van~Gool, and Radu Timofte.
\newblock Fourier space losses for efficient perceptual image super-resolution.
\newblock In \emph{Proceedings of the IEEE/CVF International Conference on Computer Vision}, pages 2360--2369, 2021.

\bibitem[Fuoli et~al.(2023)Fuoli, Danelljan, Timofte, and Van~Gool]{fuoli2023fast}
Dario Fuoli, Martin Danelljan, Radu Timofte, and Luc Van~Gool.
\newblock Fast online video super-resolution with deformable attention pyramid.
\newblock In \emph{Proceedings of the IEEE/CVF winter conference on applications of computer vision}, pages 1735--1744, 2023.

\bibitem[Gu and Dao(2023)]{gu2023mamba}
Albert Gu and Tri Dao.
\newblock Mamba: Linear-time sequence modeling with selective state spaces.
\newblock \emph{arXiv preprint arXiv:2312.00752}, 2023.

\bibitem[Gu et~al.(2021)Gu, Goel, and R{\'e}]{gu2021efficiently}
Albert Gu, Karan Goel, and Christopher R{\'e}.
\newblock Efficiently modeling long sequences with structured state spaces.
\newblock \emph{arXiv preprint arXiv:2111.00396}, 2021.

\bibitem[Guo et~al.(2024)Guo, Li, Dai, Ouyang, Ren, and Xia]{guo2024mambair}
Hang Guo, Jinmin Li, Tao Dai, Zhihao Ouyang, Xudong Ren, and Shu-Tao Xia.
\newblock Mambair: A simple baseline for image restoration with state-space model.
\newblock \emph{arXiv preprint arXiv:2402.15648}, 2024.

\bibitem[Guo et~al.(2025)Guo, Li, Dai, Ouyang, Ren, and Xia]{guo2025mambair}
Hang Guo, Jinmin Li, Tao Dai, Zhihao Ouyang, Xudong Ren, and Shu-Tao Xia.
\newblock Mambair: A simple baseline for image restoration with state-space model.
\newblock In \emph{European Conference on Computer Vision}, pages 222--241. Springer, 2025.

\bibitem[Hao and Yu(2024)]{hao2024learning}
Yukun Hao and Feihong Yu.
\newblock Learning the frequency domain aliasing for real-world super-resolution.
\newblock \emph{Electronics}, 13\penalty0 (2):\penalty0 250, 2024.

\bibitem[Hua et~al.(2022)Hua, Dai, Liu, and Le]{hua2022transformer}
Weizhe Hua, Zihang Dai, Hanxiao Liu, and Quoc Le.
\newblock Transformer quality in linear time.
\newblock In \emph{International conference on machine learning}, pages 9099--9117. PMLR, 2022.

\bibitem[Jiang et~al.(2020)Jiang, Wang, Yi, and Jiang]{jiang2020hierarchical}
Kui Jiang, Zhongyuan Wang, Peng Yi, and Junjun Jiang.
\newblock Hierarchical dense recursive network for image super-resolution.
\newblock \emph{Pattern Recognition}, 107:\penalty0 107475, 2020.

\bibitem[Jiang et~al.(2021)Jiang, Dai, Wu, and Loy]{jiang2021focal}
Liming Jiang, Bo Dai, Wayne Wu, and Chen~Change Loy.
\newblock Focal frequency loss for image reconstruction and synthesis.
\newblock In \emph{Proceedings of the IEEE/CVF international conference on computer vision}, pages 13919--13929, 2021.

\bibitem[Johnson et~al.(2016)Johnson, Alahi, and Fei-Fei]{johnson2016perceptual}
Justin Johnson, Alexandre Alahi, and Li Fei-Fei.
\newblock Perceptual losses for real-time style transfer and super-resolution.
\newblock In \emph{Computer Vision--ECCV 2016: 14th European Conference, Amsterdam, The Netherlands, October 11-14, 2016, Proceedings, Part II 14}, pages 694--711. Springer, 2016.

\bibitem[Kappeler et~al.(2016)Kappeler, Yoo, Dai, and Katsaggelos]{kappeler2016video}
Armin Kappeler, Seunghwan Yoo, Qiqin Dai, and Aggelos~K Katsaggelos.
\newblock Video super-resolution with convolutional neural networks.
\newblock \emph{IEEE transactions on computational imaging}, 2\penalty0 (2):\penalty0 109--122, 2016.

\bibitem[Kim et~al.(2016)Kim, Lee, and Lee]{kim2016accurate}
Jiwon Kim, Jung~Kwon Lee, and Kyoung~Mu Lee.
\newblock Accurate image super-resolution using very deep convolutional networks.
\newblock In \emph{Proceedings of the IEEE conference on computer vision and pattern recognition}, pages 1646--1654, 2016.

\bibitem[Kim and Cho(2023)]{kim2023whfl}
Min~Woo Kim and Nam~Ik Cho.
\newblock Whfl: Wavelet-domain high frequency loss for sketch-to-image translation.
\newblock In \emph{Proceedings of the IEEE/CVF Winter Conference on applications of computer vision}, pages 744--754, 2023.

\bibitem[Korkmaz et~al.(2024)Korkmaz, Tekalp, and Dogan]{korkmaz2024training}
Cansu Korkmaz, A~Murat Tekalp, and Zafer Dogan.
\newblock Training generative image super-resolution models by wavelet-domain losses enables better control of artifacts.
\newblock In \emph{Proceedings of the IEEE/CVF Conference on Computer Vision and Pattern Recognition}, pages 5926--5936, 2024.

\bibitem[Ledig et~al.(2017)Ledig, Theis, Husz{\'a}r, Caballero, Cunningham, Acosta, Aitken, Tejani, Totz, Wang, et~al.]{ledig2017photo}
Christian Ledig, Lucas Theis, Ferenc Husz{\'a}r, Jose Caballero, Andrew Cunningham, Alejandro Acosta, Andrew Aitken, Alykhan Tejani, Johannes Totz, Zehan Wang, et~al.
\newblock Photo-realistic single image super-resolution using a generative adversarial network.
\newblock In \emph{Proceedings of the IEEE conference on computer vision and pattern recognition}, pages 4681--4690, 2017.

\bibitem[Li et~al.(2024)Li, Li, Wang, He, Wang, Wang, and Qiao]{li2024videomamba}
Kunchang Li, Xinhao Li, Yi Wang, Yinan He, Yali Wang, Limin Wang, and Yu Qiao.
\newblock Videomamba: State space model for efficient video understanding, 2024.

\bibitem[Liang et~al.(2021)Liang, Cao, Sun, Zhang, Van~Gool, and Timofte]{liang2021swinir}
Jingyun Liang, Jiezhang Cao, Guolei Sun, Kai Zhang, Luc Van~Gool, and Radu Timofte.
\newblock Swinir: Image restoration using swin transformer.
\newblock In \emph{Proceedings of the IEEE/CVF international conference on computer vision}, pages 1833--1844, 2021.

\bibitem[Liang et~al.(2022)Liang, Fan, Xiang, Ranjan, Ilg, Green, Cao, Zhang, Timofte, and Gool]{liang2022recurrent}
Jingyun Liang, Yuchen Fan, Xiaoyu Xiang, Rakesh Ranjan, Eddy Ilg, Simon Green, Jiezhang Cao, Kai Zhang, Radu Timofte, and Luc~V Gool.
\newblock Recurrent video restoration transformer with guided deformable attention.
\newblock \emph{Advances in Neural Information Processing Systems}, 35:\penalty0 378--393, 2022.

\bibitem[Liang et~al.(2024)Liang, Cao, Fan, Zhang, Ranjan, Li, Timofte, and Van~Gool]{liang2024vrt}
Jingyun Liang, Jiezhang Cao, Yuchen Fan, Kai Zhang, Rakesh Ranjan, Yawei Li, Radu Timofte, and Luc Van~Gool.
\newblock Vrt: A video restoration transformer.
\newblock \emph{IEEE Transactions on Image Processing}, 2024.

\bibitem[Lim et~al.(2017)Lim, Son, Kim, Nah, and Mu~Lee]{lim2017enhanced}
Bee Lim, Sanghyun Son, Heewon Kim, Seungjun Nah, and Kyoung Mu~Lee.
\newblock Enhanced deep residual networks for single image super-resolution.
\newblock In \emph{Proceedings of the IEEE conference on computer vision and pattern recognition workshops}, pages 136--144, 2017.

\bibitem[Liu and Sun(2013)]{liu2013bayesian}
Ce Liu and Deqing Sun.
\newblock On bayesian adaptive video super resolution.
\newblock \emph{IEEE transactions on pattern analysis and machine intelligence}, 36\penalty0 (2):\penalty0 346--360, 2013.

\bibitem[Lu et~al.(2021)Lu, Yao, Zhang, Zhu, Xu, Gao, Xu, Xiang, and Zhang]{lu2021soft}
Jiachen Lu, Jinghan Yao, Junge Zhang, Xiatian Zhu, Hang Xu, Weiguo Gao, Chunjing Xu, Tao Xiang, and Li Zhang.
\newblock Soft: Softmax-free transformer with linear complexity.
\newblock \emph{Advances in Neural Information Processing Systems}, 34:\penalty0 21297--21309, 2021.

\bibitem[Luo et~al.(2020)Luo, Huang, and Yuan]{luo2020video}
Jianping Luo, Shaofei Huang, and Yuan Yuan.
\newblock Video super-resolution using multi-scale pyramid 3d convolutional networks.
\newblock In \emph{Proceedings of the 28th ACM International Conference on Multimedia}, pages 1882--1890, 2020.

\bibitem[Luo et~al.(2016)Luo, Li, Urtasun, and Zemel]{luo2016understanding}
Wenjie Luo, Yujia Li, Raquel Urtasun, and Richard Zemel.
\newblock Understanding the effective receptive field in deep convolutional neural networks.
\newblock \emph{Advances in neural information processing systems}, 29, 2016.

\bibitem[Mildenhall et~al.(2021)Mildenhall, Srinivasan, Tancik, Barron, Ramamoorthi, and Ng]{mildenhall2021nerf}
Ben Mildenhall, Pratul~P Srinivasan, Matthew Tancik, Jonathan~T Barron, Ravi Ramamoorthi, and Ren Ng.
\newblock Nerf: Representing scenes as neural radiance fields for view synthesis.
\newblock \emph{Communications of the ACM}, 65\penalty0 (1):\penalty0 99--106, 2021.

\bibitem[Nah et~al.(2019)Nah, Baik, Hong, Moon, Son, Timofte, and Mu~Lee]{nah2019ntire}
Seungjun Nah, Sungyong Baik, Seokil Hong, Gyeongsik Moon, Sanghyun Son, Radu Timofte, and Kyoung Mu~Lee.
\newblock Ntire 2019 challenge on video deblurring and super-resolution: Dataset and study.
\newblock In \emph{Proceedings of the IEEE/CVF conference on computer vision and pattern recognition workshops}, pages 0--0, 2019.

\bibitem[Nasiri-Sarvi et~al.(2024)Nasiri-Sarvi, Trinh, Rivaz, and Hosseini]{nasiri2024vim4path}
Ali Nasiri-Sarvi, Vincent Quoc-Huy Trinh, Hassan Rivaz, and Mahdi~S Hosseini.
\newblock Vim4path: Self-supervised vision mamba for histopathology images.
\newblock In \emph{Proceedings of the IEEE/CVF Conference on Computer Vision and Pattern Recognition}, pages 6894--6903, 2024.

\bibitem[Park et~al.(2018)Park, Son, Cho, Hong, and Lee]{park2018srfeat}
Seong-Jin Park, Hyeongseok Son, Sunghyun Cho, Ki-Sang Hong, and Seungyong Lee.
\newblock Srfeat: Single image super-resolution with feature discrimination.
\newblock In \emph{Proceedings of the European conference on computer vision (ECCV)}, pages 439--455, 2018.

\bibitem[Rad et~al.(2019)Rad, Bozorgtabar, Marti, Basler, Ekenel, and Thiran]{rad2019srobb}
Mohammad~Saeed Rad, Behzad Bozorgtabar, Urs-Viktor Marti, Max Basler, Hazim~Kemal Ekenel, and Jean-Philippe Thiran.
\newblock Srobb: Targeted perceptual loss for single image super-resolution.
\newblock In \emph{Proceedings of the IEEE/CVF international conference on computer vision}, pages 2710--2719, 2019.

\bibitem[Rahaman et~al.(2019)Rahaman, Baratin, Arpit, Draxler, Lin, Hamprecht, Bengio, and Courville]{rahaman2019spectral}
Nasim Rahaman, Aristide Baratin, Devansh Arpit, Felix Draxler, Min Lin, Fred Hamprecht, Yoshua Bengio, and Aaron Courville.
\newblock On the spectral bias of neural networks.
\newblock In \emph{International conference on machine learning}, pages 5301--5310. PMLR, 2019.

\bibitem[Ranjan and Black(2016)]{ranjan2016opticalflowestimationusing}
Anurag Ranjan and Michael~J. Black.
\newblock Optical flow estimation using a spatial pyramid network, 2016.

\bibitem[Shi et~al.(2022)Shi, Gu, Xie, Wang, Yang, and Dong]{shi2022rethinking}
Shuwei Shi, Jinjin Gu, Liangbin Xie, Xintao Wang, Yujiu Yang, and Chao Dong.
\newblock Rethinking alignment in video super-resolution transformers.
\newblock \emph{Advances in Neural Information Processing Systems}, 35:\penalty0 36081--36093, 2022.

\bibitem[Smith et~al.(2023)Smith, Warrington, and Linderman]{smith2023simplified}
Jimmy~T.H. Smith, Andrew Warrington, and Scott Linderman.
\newblock Simplified state space layers for sequence modeling.
\newblock In \emph{The Eleventh International Conference on Learning Representations}, 2023.

\bibitem[Su et~al.(2019)Su, Jampani, Sun, Gallo, Learned-Miller, and Kautz]{su2019pixel}
Hang Su, Varun Jampani, Deqing Sun, Orazio Gallo, Erik Learned-Miller, and Jan Kautz.
\newblock Pixel-adaptive convolutional neural networks.
\newblock In \emph{Proceedings of the IEEE/CVF Conference on Computer Vision and Pattern Recognition}, pages 11166--11175, 2019.

\bibitem[Tancik et~al.(2020)Tancik, Srinivasan, Mildenhall, Fridovich-Keil, Raghavan, Singhal, Ramamoorthi, Barron, and Ng]{tancik2020fourier}
Matthew Tancik, Pratul Srinivasan, Ben Mildenhall, Sara Fridovich-Keil, Nithin Raghavan, Utkarsh Singhal, Ravi Ramamoorthi, Jonathan Barron, and Ren Ng.
\newblock Fourier features let networks learn high frequency functions in low dimensional domains.
\newblock \emph{Advances in neural information processing systems}, 33:\penalty0 7537--7547, 2020.

\bibitem[Tian et~al.(2020)Tian, Zhang, Fu, and Xu]{tian2020tdan}
Yapeng Tian, Yulun Zhang, Yun Fu, and Chenliang Xu.
\newblock Tdan: Temporally-deformable alignment network for video super-resolution.
\newblock In \emph{Proceedings of the IEEE/CVF conference on computer vision and pattern recognition}, pages 3360--3369, 2020.

\bibitem[Tran et~al.(2022)Tran, Nguyen, Pham, and Tran]{tran2022trans2unet}
Dinh-Phu Tran, Quoc-Anh Nguyen, Van-Truong Pham, and Thi-Thao Tran.
\newblock Trans2unet: neural fusion for nuclei semantic segmentation.
\newblock In \emph{2022 11th international conference on control, automation and information sciences (ICCAIS)}, pages 583--588. IEEE, 2022.

\bibitem[Tran et~al.(2024)Tran, Hung, and Kim]{tran2024channel}
Dinh~Phu Tran, Dao~Duy Hung, and Daeyoung Kim.
\newblock Channel-partitioned windowed attention and frequency learning for single image super-resolution.
\newblock \emph{arXiv preprint arXiv:2407.16232}, 2024.

\bibitem[Vaswani(2017)]{vaswani2017attention}
A Vaswani.
\newblock Attention is all you need.
\newblock \emph{Advances in Neural Information Processing Systems}, 2017.

\bibitem[Wang et~al.(2019)Wang, Chan, Yu, Dong, and Change~Loy]{wang2019edvr}
Xintao Wang, Kelvin~CK Chan, Ke Yu, Chao Dong, and Chen Change~Loy.
\newblock Edvr: Video restoration with enhanced deformable convolutional networks.
\newblock In \emph{Proceedings of the IEEE/CVF conference on computer vision and pattern recognition workshops}, pages 0--0, 2019.

\bibitem[Wang et~al.(2004)Wang, Bovik, Sheikh, and Simoncelli]{wang2004image}
Zhou Wang, Alan~C Bovik, Hamid~R Sheikh, and Eero~P Simoncelli.
\newblock Image quality assessment: from error visibility to structural similarity.
\newblock \emph{IEEE transactions on image processing}, 13\penalty0 (4):\penalty0 600--612, 2004.

\bibitem[Xu et~al.(2024)Xu, Yu, Wang, Mi, and Yao]{xu2024enhancing}
Kai Xu, Ziwei Yu, Xin Wang, Michael~Bi Mi, and Angela Yao.
\newblock Enhancing video super-resolution via implicit resampling-based alignment.
\newblock In \emph{Proceedings of the IEEE/CVF Conference on Computer Vision and Pattern Recognition}, pages 2546--2555, 2024.

\bibitem[Xue et~al.(2019)Xue, Chen, Wu, Wei, and Freeman]{xue2019video}
Tianfan Xue, Baian Chen, Jiajun Wu, Donglai Wei, and William~T Freeman.
\newblock Video enhancement with task-oriented flow.
\newblock \emph{International Journal of Computer Vision}, 127:\penalty0 1106--1125, 2019.

\bibitem[Yang et~al.(2021)Yang, Xiang, Zeng, and Zhang]{yang2021real}
Xi Yang, Wangmeng Xiang, Hui Zeng, and Lei Zhang.
\newblock Real-world video super-resolution: A benchmark dataset and a decomposition based learning scheme.
\newblock In \emph{Proceedings of the IEEE/CVF International Conference on Computer Vision}, pages 4781--4790, 2021.

\bibitem[Yang et~al.(2024)Yang, Xing, and Zhu]{yang2024vivim}
Yijun Yang, Zhaohu Xing, and Lei Zhu.
\newblock Vivim: a video vision mamba for medical video object segmentation.
\newblock \emph{arXiv preprint arXiv:2401.14168}, 2024.

\bibitem[Zhou et~al.(2024)Zhou, Zhang, Zhao, Wang, Li, and Gu]{zhou2024video}
Xingyu Zhou, Leheng Zhang, Xiaorui Zhao, Keze Wang, Leida Li, and Shuhang Gu.
\newblock Video super-resolution transformer with masked inter\&intra-frame attention.
\newblock In \emph{Proceedings of the IEEE/CVF Conference on Computer Vision and Pattern Recognition}, pages 25399--25408, 2024.

\bibitem[Zhu et~al.(2024)Zhu, Liao, Zhang, Wang, Liu, and Wang]{vim2024}
Lianghui Zhu, Bencheng Liao, Qian Zhang, Xinlong Wang, Wenyu Liu, and Xinggang Wang.
\newblock Vision mamba: Efficient visual representation learning with bidirectional state space model.
\newblock In \emph{Forty-first International Conference on Machine Learning}, 2024.

\end{thebibliography}
}


\end{document}